\documentclass[letterpaper, 10 pt, conference]{ieeeconf}  

\IEEEoverridecommandlockouts                              

\usepackage{hyperref}
\usepackage{url}
\usepackage{cite}
\usepackage{graphicx}
\usepackage{gensymb}
\usepackage{amsmath}
\usepackage{tabularx}
\usepackage{color,soul}

\title{\LARGE \bf
Supersizing Self-supervision: Learning to Grasp \\ from 50K Tries and 700 Robot Hours
}
\author{Lerrel Pinto and Abhinav Gupta
\\ The Robotics Institute, Carnegie Mellon University
\\ \texttt{(lerrelp, abhinavg)@cs.cmu.edu}
\vspace*{-0.1in}
}

\begin{document}
\maketitle

\thispagestyle{empty}
\pagestyle{empty}


\begin{abstract}
Current learning-based robot grasping approaches exploit human-labeled datasets for training the models. However, there are two problems with such a methodology: (a) since each object can be grasped in multiple ways, manually labeling grasp locations is not a trivial task; (b) human labeling is biased by semantics. While there have been attempts to train robots using trial-and-error experiments, the amount of data used in such experiments remains substantially low and hence makes the learner prone to over-fitting. In this paper, we take the leap of increasing the available training data to 40 times more than prior work, leading to a dataset size of 50K data points collected over 700 hours of robot grasping attempts. This allows us to train a Convolutional Neural Network (CNN) for the task of predicting grasp locations without severe overfitting. In our formulation, we recast the regression problem to an 18-way binary classification over image patches. We also present a multi-stage learning approach where a CNN trained in one stage is used to collect hard negatives in subsequent stages. Our experiments clearly show the benefit of using large-scale datasets (and multi-stage training) for the task of grasping. We also compare to several baselines and show state-of-the-art performance on generalization to unseen objects for grasping.
\end{abstract}

\section{INTRODUCTION}
\begin{figure}[t!]
\begin{center}
\includegraphics[width=3.5in]{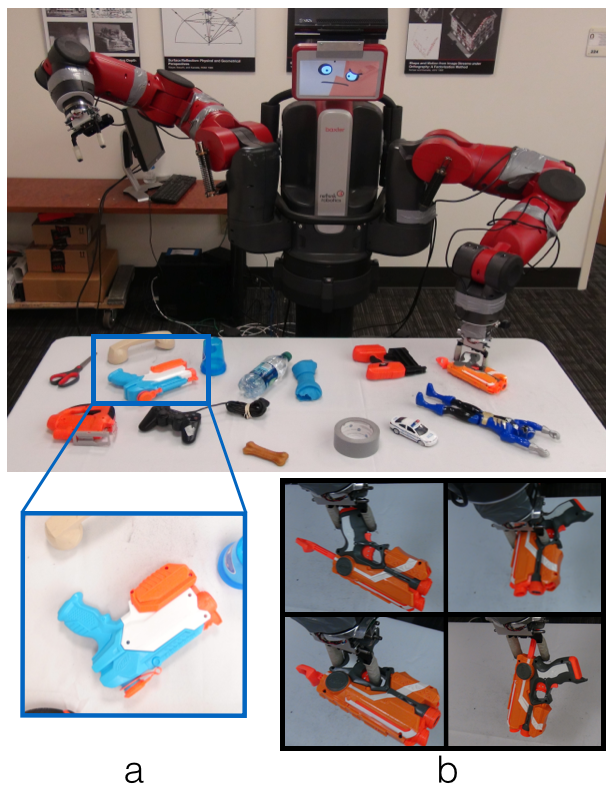}
\end{center}
\vspace{-0.1in}
\caption{ We present an approach to train robot grasping using 50K trial and error grasps. Some of the sample objects and our setup are shown in (a). Note that each object in the dataset can be grasped in multiple ways (b) and therefore exhaustive human labeling of this task is extremely difficult.}
\vspace{-0.2in}
\label{fig:intro_fig}
\end{figure}

Consider the object shown in Fig.~\ref{fig:intro_fig}(a). How do we predict grasp locations for this object? One approach is to fit 3D models to these objects, or to use a 3D depth sensor, and perform analytical 3D reasoning to predict the grasp locations~\cite{brooks1983planning,shimoga1996robot,lozano1989task,nguyen1988constructing}. However, such an approach has two drawbacks: (a) fitting 3D models is an extremely difficult problem by itself; but more importantly, (b) a geometry based-approach ignores the densities and mass distribution of the object which may be vital in predicting the grasp locations. Therefore, a more practical approach is to use visual recognition to predict grasp locations and configurations, since it does not require explicit modelling of objects. For example, one can create a grasp location training dataset for hundreds and thousands of objects and use standard machine learning algorithms such as CNNs~\cite{le1990handwritten,krizhevsky2012imagenet} or autoencoders~\cite{olshausen1997sparse} to predict grasp locations in the test data. However, creating a grasp dataset using human labeling can itself be quite challenging for two reasons. First, most objects can be grasped in multiple ways which makes exhaustive labeling impossible (and hence negative data is hard to get; see Fig.~\ref{fig:intro_fig}(b)). Second, human notions of grasping are biased by semantics. For example, humans tend to label handles as the grasp location for objects like cups even though they might be graspable from several other locations and configurations. Hence, a randomly sampled patch cannot be assumed to be a negative data point, even if it was not marked as a positive grasp location by a human. Due to these challenges, even the biggest vision-based grasping dataset~\cite{jiang2011efficient} has about only 1K images of objects in isolation (only {\bf one} object visible without any clutter).

In this paper, we break the mold of using manually labeled grasp datasets for training grasp models. We believe such an approach is not scalable. Instead, inspired by reinforcement learning (and human experiential learning), we present a self-supervising algorithm that learns to predict grasp locations via trial and error. But how much training data do we need to train high capacity models such as Convolutional Neural Networks (CNNs)~\cite{krizhevsky2012imagenet} to predict meaningful grasp locations for new unseen objects? Recent approaches have tried to use reinforcement learning with a few hundred datapoints and learn a CNN with hundreds of thousand parameters~\cite{levine2015end}. We believe that such an approach, where the training data is substantially fewer than the number of model parameters, is bound to overfit and would fail to generalize to new unseen objects. Therefore, what we need is a way to collect hundreds and thousands of data points (possibly by having a robot interact with objects 24/7) to learn a meaningful representation for this task. But is it really possible to scale trial and errors experiments to learn visual representations for the task of grasp prediction?

\begin{figure*}[t!]
\begin{center}
\includegraphics[width=7in]{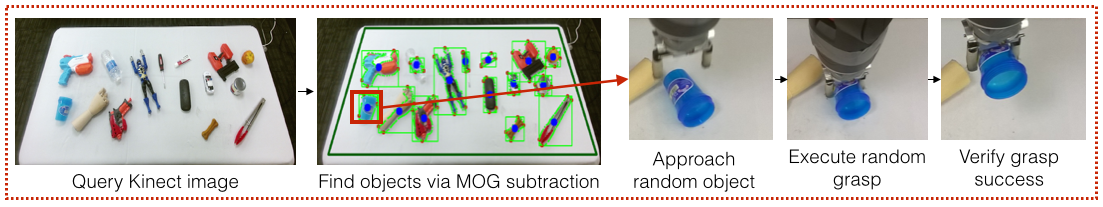}
\end{center}
\vspace{-0.1in}
\caption{Overview of how random grasp actions are sampled and executed.
}
\label{fig:data_collection_method}
\vspace{-0.1in}
\end{figure*}

 Given the success of high-capacity learning algorithms such as CNNs, we believe it is time to develop large-scale robot datasets for foundational tasks such as grasping. Therefore, we present a large-scale experimental study that not only substantially increases the amount of data for learning to grasp, but provides complete labeling in terms of whether an object can be grasped at a particular location and angle. This dataset, collected with robot executed interactions, will be released for research use to the community. We use this dataset to fine-tune an AlexNet~\cite{krizhevsky2012imagenet} CNN model pre-trained on ImageNet, with 18M new parameters to learn in the fully connected layers, for the task of prediction of grasp location. Instead of using regression loss, we formulate the problem of grasping as an 18-way binary classification over 18 angle bins. Inspired by the reinforcement learning paradigm~\cite{ross2010reduction,mnih2015human}, we also present a staged-curriculum based learning algorithm where we learn how to grasp, and use the most recently learned model to collect more data.

The contributions of the paper are three-fold: (a) we introduce one of the largest robot datasets for the task of grasping. Our dataset has more than 50K datapoints and has been collected using 700 hours of trial and error experiments using the Baxter robot. (b) We present a novel formulation of CNN for the task of grasping. We predict grasping locations by sampling image patches and predicting the grasping angle. Note that since an object may be graspable at multiple angles, we model the output layer as an 18-way binary classifier. (c) We present a multi-stage learning approach to collect hard-negatives and learn a better grasping model. Our experiments clearly indicate that a larger amount of data is helpful in learning a better grasping model. We also show the importance of multi-stage learning using ablation studies and compare our approach to several baselines. Real robot testing is performed to validate our method and show generalization to grasping unseen objects.
\section{Related Work}

Object manipulation is one of the oldest problems in the field of robotics. A comprehensive literature review of this area can be found in~\cite{bicchi2000robotic,bohg2014data}. Early attempts in the field focused on using analytical methods and 3D reasoning for predicting grasp locations and configurations~\cite{brooks1983planning,shimoga1996robot,lozano1989task,nguyen1988constructing}. These approaches assumed the availability of complete knowledge of the objects to be grasped, such as the complete 3D model of the given object, along with the object's surface friction properties and mass distribution. However, perception and inference of 3D models and other attributes such as friction/mass from RGB or RGBD cameras is an extremely difficult problem. To solve these problems people have constructed grasp databases \cite{goldfeder2009columbia,kootstra2012visgrab}. Grasps are sampled and ranked based on similarities to grasp instances in a pre-existing database. These methods however do not generalize well to objects outside the database.

Other approaches to predict grasping includes using simulators such as Graspit!\cite{miller2004graspit,miller2003automatic}. In these approaches, one samples grasp candidates and ranks them based on an analytical formulation. However questions often arise as to how well a simulated environment mirrors the real world. \cite{bohg2014data,diankov2010automated,weisz2012pose} offer reasons as to why a simulated environment and an analytic metric would not parallel the real world which is highly unstructured.

Recently, there has been more focus on using visual learning to predict grasp locations directly from RGB or RGB-D images~\cite{saxena2008robotic,montesano2012active} . For example, \cite{saxena2008robotic} uses vision based features (edge and texture filter responses) and learns a logistic regressor over synthetic data. On the other hand, \cite{lenz2013deep,ramisa2012using} use human annotated grasp data to train grasp synthesis models over RGB-D data. However, as discussed above, large-scale collection of training data for the task of grasp prediction is not trivial and has several issues. Therefore, none of the above approaches are scalable to use big data.

Another common way to collect data for robotic tasks is using the robot's own trial and error experiences ~\cite{morales2004using,detry2009learning,Paolini_2014_7585}. However, even recent approaches such as~\cite{levine2015end,levine2015learning} only use a few hundred trial and error runs to train high capacity deep networks. We believe this causes the network to overfit and often no results are shown on generalizability to new unseen objects. Other approaches in this domain such as \cite{BoulariasBS15} use reinforcement learning to learn grasp attributes over depth images of a cluttered scene. However the grasp attributes are based on supervoxel segmentation and facet detection. This creates a prior on grasp synthesis and may not be desirable for complex objects.  

Deep neural networks have seen immense success in image classification \cite{krizhevsky2012imagenet} and object detection \cite{girshick2014rich}. Deep networks have also been exploited in robotics systems for grasp regression \cite{redmon2014real} or learning policy for variety of tasks~\cite{levine2015learning}. Furthermore DAgger~\cite{ross2010reduction} shows a simple and practical method of sampling the interesting regions of a state space by dataset aggregation. In this paper, we propose an approach to scale up the learning from few hundred examples to thousands of examples. We present an end-to-end self-supervising staged curriculum learning system that uses thousands of trial-error runs to learn deep networks. The learned deep network is then used to collect greater amounts of positive and hard negative (model thinks as graspable but in general are not) data which helps the network to learn faster. 
\section{Approach}

We first explain our robotic grasping system and how we use it to collect more than 50K data points. 
Given these training data points, we train a CNN-based classifier which given an input image patch predicts the grasp likelihood for different grasp directions. Finally, we explain our staged-curriculum learning framework which helps our system to find hard negatives: data points on which the model performs poorly and hence causes high loss with greater back propagation signal. 

\noindent {\bf Robot Grasping System:} Our experiments are carried out on a Baxter robot from Rethink Robotics and we use ROS \cite{quigley2009ros} as our development system. For gripping we use the stock two fingered parallel gripper with a maximum width (open state) of $75$mm and a minimum width (close state) of $37$mm.

A Kinect V2 is attached to the head of the robot that provides $1920\times1280$ resolution image of the workspace(dull white colored table-top). Furthermore, a $1280\times720$ resolution camera is attached onto each of Baxter's end effector which provides rich images of the objects Baxter interacts with. For the purposes of trajectory planning a stock Expansive Space Tree (EST) planner \cite{sucan2012the-open-motion-planning-library} is used. It should be noted that we use both the robot arms to collect the data more quickly.

During experiments, human involvement is limited to switching on the robot and placing the objects on the table in an arbitrary manner. Apart from initialization, we have {\bf no human involvement} in the process of data collection. Also, in order to gather data as close to real world test conditions, we perform trial and error grasping experiments in cluttered environment. 
Grasped objects, on being dropped, at times bounce/roll off the robot workspace, however using cluttered environments also ensures that the robot always has an object to grasp. This experimental setup negates the need for constant human supervision. The Baxter robot is also robust against break down, with experiments running for 8-10 hours a day.

\noindent {\bf Gripper Configuration Space and Parametrization:} In this paper, we focus on the planar grasps only.  A planar grasp is one where the grasp configuration is along and perpendicular to the workspace. Hence the grasp configuration lies in 3 dimensions, $(x,y)$: position of grasp point on the surface of table and  $\theta$: angle of grasp.

\subsection{Trial and Error Experiments}

The data collection methodology is succinctly described in Fig.~\ref{fig:data_collection_method}. The workspace is first setup with multiple objects of varying difficulty of graspability placed haphazardly on a table with a dull white background. Multiple random trials are then executed in succession. 

A single instance of a random trial goes as follows:

\textbf{Region of Interest Sampling:} An image of the table, queried from the head-mounted Kinect, is passed through an off-the-shelf Mixture of Gaussians (MOG) background subtraction algorithm that identifies regions of interest in the image. This is done solely to reduce the number of random trials in empty spaces without objects in the vicinity. A random region in this image is then selected to be the region of interest for the specific trial instance.

\textbf{Grasp Configuration Sampling:} Given a specific region of interest, the robot arm moves to $25$cm above the object. Now a random point is uniformly sampled from the space in the region of interest. This will be the robot's grasp point. To complete the grasp configuration, an angle is now chosen randomly in range$(0,\pi)$ since the two fingered gripper is symmetric.

\textbf{Grasp Execution and Annotation:} Now given the grasp configuration, the robot arm executes a pick grasp on the object. The object is then raised by $20$cm and annotated as a success or a failure depending on the gripper's force sensor readings.

Images from all the cameras, robot arm trajectories and gripping history are recorded to disk during the execution of these random trials.

\begin{figure}[t!]
\begin{center}
\includegraphics[width=3.4in]{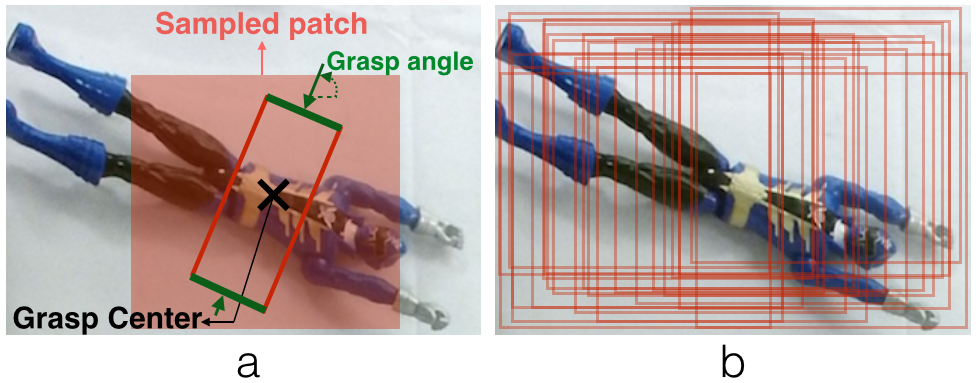}
\end{center}
\vspace{-0.1in}
\caption{(a) We use 1.5 times the gripper size image patch to predict the grasp-ability of a location and the angle at which it can be grasped. Visualization for showing the grasp location and the angle of gripper for grasping is derived from \cite{jiang2011efficient}. (b) At test time we sample patches at different positions and choose the top graspable location and corresponding gripper angle.}
\vspace{-0.1in}
\label{fig:grasp_config}
\end{figure}
\subsection{Problem Formulation}
\begin{figure*}[t!]
\begin{center}
\includegraphics[width=7in]{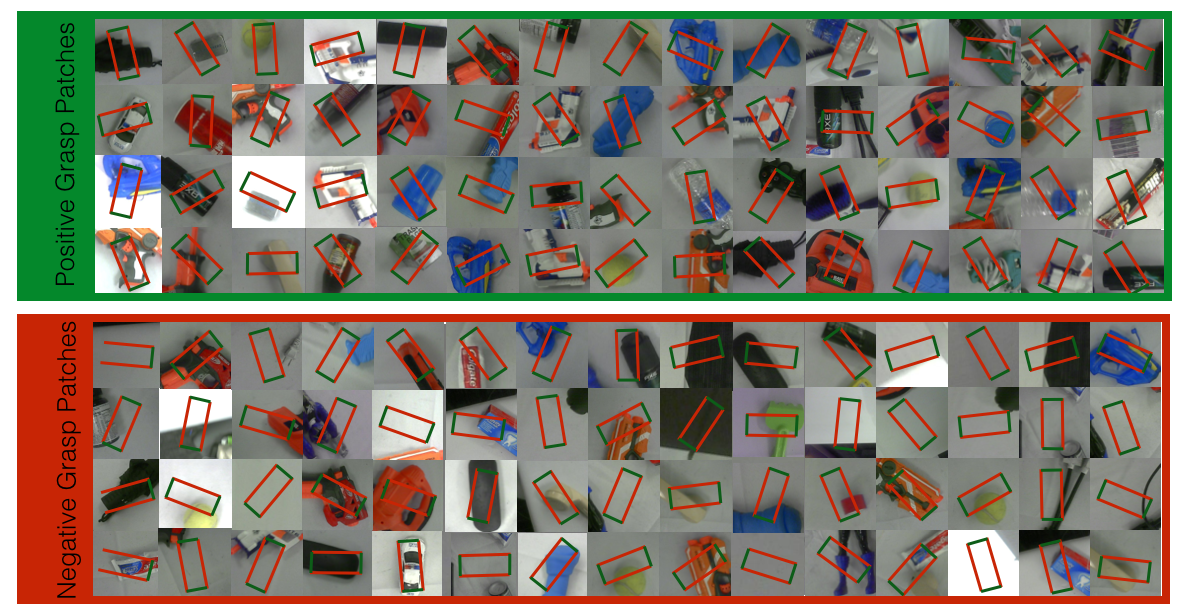}
\end{center}
\vspace{-0.15in}
\caption{Sample patches used for training the Convolutional Neural Network.}
\label{fig:training_data}
\end{figure*}

The grasp synthesis problem is formulated as finding a successful grasp configuration $(x_S,y_S,\theta_S)$ given an image of an object $I$. 
A grasp on the object can be visualised using the rectangle representation \cite{jiang2011efficient} in Fig.~\ref{fig:grasp_config}. In this paper, we use CNNs to predict grasp locations and angle. We now explain the input and output to the CNN.

\noindent {\bf Input:} The input to our CNN is an image patch extracted around the grasp point. For our experiments, we use patches 1.5 times as large as the projection of gripper fingertips on the image, to include context as well. The patch size used in experiments is 380x380. This patch is resized to 227x227 which is the input image size of the ImageNet-trained AlexNet~\cite{krizhevsky2012imagenet}.

\noindent {\bf Output:}  One can train the grasping problem as a regression problem: that is, given an input image predict $(x,y,\theta)$. However, this formulation is problematic since: (a) there are multiple grasp locations for each object; (b) CNNs are significantly better at classification than the regressing to a structured output space. Another possibility is to formulate this as a two-step classification: that is, first learn a binary classifier model that classifies the patch as graspable or not and then selects the grasp angle for positive patches. However graspability of an image patch is a function of the angle of the gripper, and therefore an image patch can be labeled as both graspable and non-graspable.

Instead, in our case, given an image patch we estimate an 18-dimensional likelihood vector where each dimension represents the likelihood of whether the center of the patch is graspable at $0^{\circ}$, $10^{\circ}$, \dots $170^{\circ}$. Therefore, our problem can be thought of an 18-way binary classification problem.

\noindent {\bf Testing:} Given an image patch, our CNN outputs whether an object is graspable at the center of the patch for the 18 grasping angles. At test time on the robot, given an image, we sample grasp locations and extract patches which is fed into the CNN. For each patch, the output is 18 values which depict the graspability scores for each of the 18 angles. We select the maximum score across all angles and all patches, and execute grasp at the corresponding grasp location and angle. 

\subsection{Training Approach}
\noindent{\bf Data preparation:} Given a trial experiment datapoint $(x_i, y_i, \theta_i)$, we sample 380x380 patch with $(x_i,y_i)$ being the center. To increase the amount of data seen by the network, we use rotation transformations: rotate the dataset patches by $\theta_{rand}$ and label the corresponding grasp orientation as $\{\theta_{i} + \theta_{rand}\}$. Some of these patches can be seen in Fig.~\ref{fig:training_data}

\begin{figure*}[t!]
\begin{center}
\includegraphics[trim=0.2in 4.5in 0.4in 2in, clip=true, width=7in]{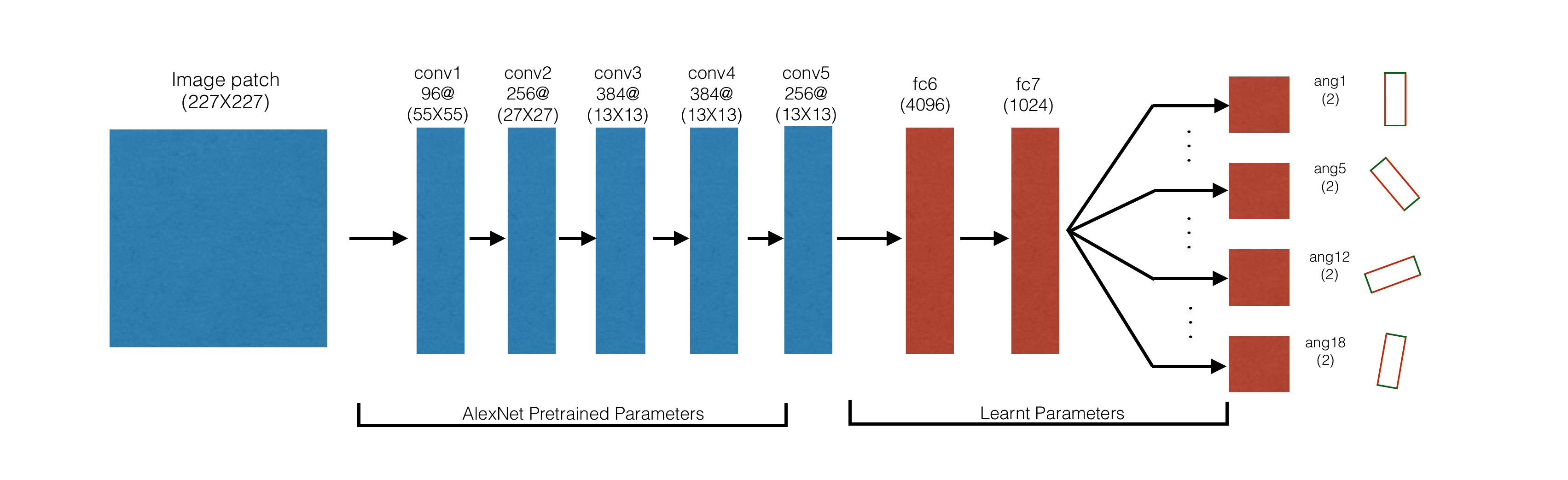}
\end{center}
\vspace{-0.3in}
\caption{Our CNN architecture is similar to AlexNet~\cite{krizhevsky2012imagenet}. We initialize our convolutional layers from ImageNet-trained Alexnet.}
\vspace{-0.1in}
\label{fig:network}
\end{figure*}

\noindent {\bf Network Design:} Our CNN, seen in Fig.~\ref{fig:network}, is a standard network architecture: our first five convolutional layers are taken from the AlexNet~\cite{krizhevsky2012imagenet,jia2014caffe} pretrained on ImageNet.  We also use two fully connected layers with 4096 and 1024 neurons respectively. The two fully connected layers, fc6 and fc7 are trained with gaussian initialisation.

\noindent{\bf Loss Function:} The loss of the network is formalized as follows. Given a batch size $B$, with a patch instance $P_i$, let the label corresponding to angle $\theta_i$ be defined by $l_i\in \{0,1\}$ and the forward pass binary activations $A_{ji}$ (vector of length 2) on the angle bin $j$ \, we define our batch loss $L_B$ as:

\begin{equation}
L_B = \sum\limits_{i=1}^B\sum\limits_{j=1}^{N=18}\delta(j,\theta_i)\cdotp \textrm{softmax}(A_{ji},\l_i)
\end{equation}

where, $\delta(j,\theta_i) = 1$ when $\theta_i$ corresponds to $j^{th}$ bin. Note that the last layer of the network involves 18 binary layers instead of one multiclass layer to predict the final graspability scores. Therefore, for a single patch, only the loss corresponding to the trial angle bin is backpropagated. 

\subsection{Staged Learning}
\begin{figure}[t!]
\begin{center}
\includegraphics[width=3.4in]{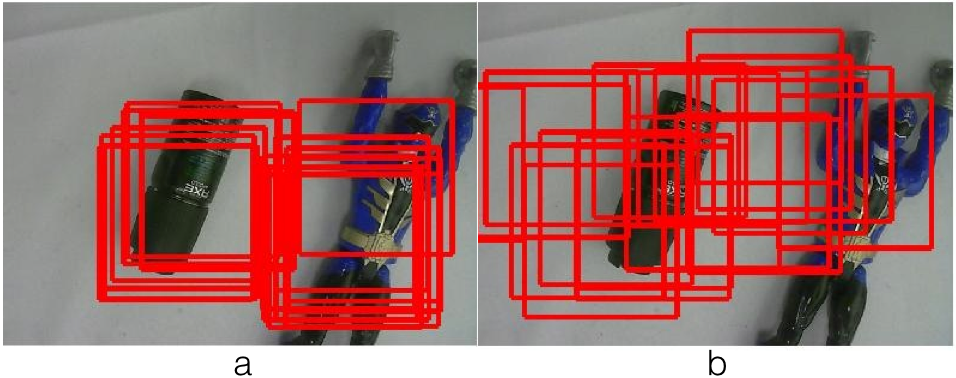}
\end{center}
\vspace{-0.15in}
\caption{Highly ranked patches from learnt algorithm (a) focus more on the objects in comparison to random patches (b).}
\vspace{-0.2in}
\label{fig:patch_proposal}
\end{figure}

Given the network trained on the random trial experience dataset, the robot now uses this model as a prior on grasping. At this stage of data collection, we use both previously seen objects and novel objects. This ensures that in the next iteration, the robot corrects for incorrect grasp modalities while reinforcing the correct ones. Fig.~\ref{fig:patch_proposal} shows how top ranked patches from a learned model focus more on important regions of the image compared to random patches. Using novel objects further enriches the model and avoids over-fitting.

Note that for every trial of object grasp at this stage, 800 patches are randomly sampled and evaluated by the deep network learnt in the previous iteration. This produces a $800\times18$ grasp-ability prior matrix where entry ($i,j$) corresponds to the network activation on the $j^{th}$ angle bin for the $i^{th}$ patch. Grasp execution is now decided by importance sampling over the grasp-ability prior matrix.

Inspired by data aggregation techniques\cite{ross2010reduction}, during training of iteration $k$, the dataset $D_k$ is given by $\{D_k\} = \{D_{k-1},\Gamma d_{k}\}$, where $d_{k}$ is the data collected using the model from iteration $k-1$. Note that $D_0$ is the random grasp dataset and iteration $0$ is simply trained on $D_0$. The importance factor $\Gamma$ is kept at 3 as a design choice.

The deep network to be used for the $k^{th}$ stage is trained by finetuning the previously trained network with dataset $D_k$. Learning rate for iteration $0$ is chosen as 0.01 and trained over 20 epochs. The remaining iterations are trained with a learning rate of 0.001 over 5 epochs.



\section{Results}
\subsection{Training dataset}
\begin{figure}[t!]
\begin{center}
\includegraphics[width=3in]{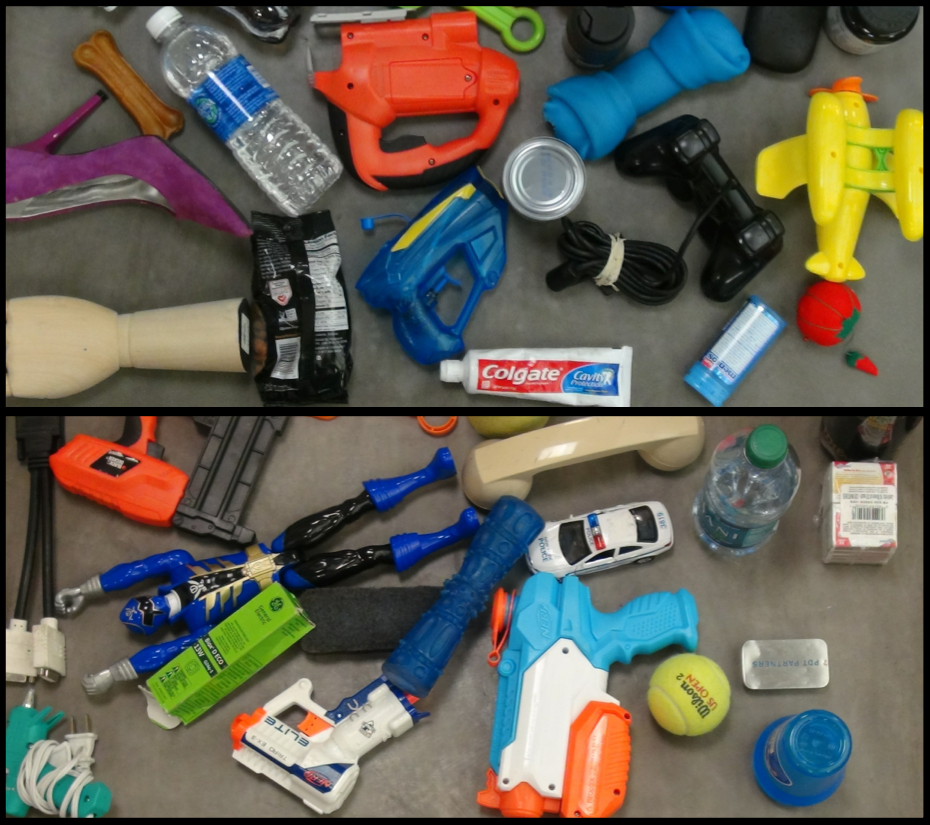}
\end{center}
\vspace{-0.15in}
\caption{Random Grasp Sampling Scenario: Our data is collected in clutter rather than objects in isolation. This allows us to generalize and tackle tasks like clutter removal.}
\vspace{-0.15in}
\label{fig:table_top}
\end{figure}

The training dataset is collected over 150 objects with varying graspability. A subset of these objects can be seen in Fig.~\ref{fig:table_top}. At the time of data collection, we use a cluttered table rather than objects in isolation. Through our large data collection and learning approach, we collect 50K grasp experience interactions. A brief summary of the data statistics can be found in Table \ref{tab:grasp_data_stat}. 

\begin{table}[h]
\caption{Grasp Dataset Statistics}
\label{tab:grasp_data_stat}
\centering
\begin{tabular}{l|c|c|c|c}
\hline
\multicolumn{1}{|l|}{{\bf \begin{tabular}{@{}c@{}}Data Collection \\ Type\end{tabular}}} & {\bf Positive} & {\bf Negative} & {\bf Total }  & \multicolumn{1}{c|}{{\bf Grasp Rate}} \\ \hline
\multicolumn{1}{|l|}{Random Trials}                   & 3,245                & 37,042               & 40,287       & \multicolumn{1}{c|}{8.05\%}            \\ \hline
\multicolumn{1}{|l|}{Multi-Staged}           & 2,807               & 4,500                & 7,307        & \multicolumn{1}{c|}{38.41\%}           \\ \hline
\multicolumn{1}{|l|}{Test Set}                   & 214                & 2,759               & 2,973       & \multicolumn{1}{c|}{7.19\%}            \\ \hline
\multicolumn{1}{c|}{}                                 & {\bf 6,266}          & {\bf 44,301}         & {\bf 50,567} &                                       \\ \cline{2-4}
\end{tabular}
\end{table}

\subsection{Testing and evaluation setting}
For comparisons with baselines and to understand the relative importance of the various components in our learning method, we report results on a held out test set with objects not seen in the training (Fig.~\ref{fig:novel_test_objects}). Grasps in the test set are collected via 3K physical robot interactions on 15 novel and diverse test objects in multiple poses. Note that this test set is balanced by random sampling from the collected robot interactions. The accuracy measure used to evaluate is binary classification i.e. given a patch and executed grasp angle in the test set, to predict whether the object was grasped or not.

Evaluation by this method preserves two important aspects for grasping: (a) It ensures that the test data is exactly the same for comparison which isn't possible with real robot experiments. (b) The data is from a real robot which means methods that work well on this test set should work well on the real robot. Our deep learning based approach followed by multi-stage reinforcement yields an accuracy of \textbf{79.5\%} on this test set. A summary of the baselines can be seen in Table. \ref{Tab:baseline_comp}.

We finally demonstrate evaluation in the real robot settings for grasping objects in isolation and show results on clearing a clutter of objects.

\subsection{Comparison with heuristic baselines} 

\begin{table*}[]
\centering
\caption{Comparing our method with baselines}
\label{Tab:baseline_comp}
\begin{tabular}{cccc|cccc}
\hline
         & \multicolumn{3}{c|}{Heuristic}                                                                                                                                                                & \multicolumn{4}{c}{Learning based}                                                                                                                  \\ \hline
         & \begin{tabular}[c]{@{}c@{}}Min \\ eigenvalue\end{tabular} & \begin{tabular}[c]{@{}c@{}}Eigenvalue \\ limit\end{tabular} & \begin{tabular}[c]{@{}c@{}}Optimistic \\ param. select\end{tabular} & kNN   & SVM   & \begin{tabular}[c]{@{}c@{}}Deep Net\\ (ours)\end{tabular} & \begin{tabular}[c]{@{}c@{}}Deep Net + Multi-stage\\ (ours)\end{tabular} \\ \hline
Accuracy & 0.534                                                     & 0.599                                                       & 0.621                                                               & 0.694 & 0.733 & 0.769                                                     & \textbf{0.795}                                                                  
\end{tabular}
\end{table*}

A strong baseline is the "common-sense" heuristic which is discussed in \cite{katz2014perceiving}. The heuristic, modified for the RGB image input task, encodes obvious grasping rules:
\begin{enumerate}
  \item Grasp about the center of the patch. This rule is implicit in our formulation of patch based grasping.
  \item Grasp about the smallest object width. This is implemented via object segmentation followed by eigenvector analysis. Heuristic's optimal grasp is chosen along the direction of the smallest eigenvalue. If the test set executed successful grasp is within an error threshold of the heuristic grasp, the prediction is a success. This leads to an accuracy of 53.4\%
  \item Do not grasp too thin objects, since the gripper doesn't close completely. If the largest eigenvalue is smaller than the mapping of the gripper's minimum width in image space, the heuristic predicts no viable grasps; i.e no object is large enough to be grasped. This leads to an accuracy of 59.9\%
\end{enumerate}
By iterating over all possible parameters (error thresholds and eigenvalue limits) in the above heuristic over the test set, the maximal accuracy obtained was 62.11\% which is significantly lower than our method's accuracy. The low accuracy is understandable since the heuristic doesn't work well for objects in clutter.

\subsection{Comparison with learning based baselines}
We now compare with a couple of learning based algorithms. We use HoG\cite{dalal2005histograms} features in both the following baselines since it preserves rotational variance which is important to grasping:
\begin{enumerate}
    \item k Nearest Neighbours (kNN): For every element in the test set, kNN based classification is performed over elements in the train set that belong to the same angle class. Maximal accuracy over varying k (optimistic kNN) is 69.4\%.
    \item Linear SVM: 18 binary SVMs are learnt for each of the 18 angle bins. After choosing regularisation parameters via validation, the maximal accuracy obtained is 73.3\%
\end{enumerate}

\subsection{Ablative analysis}

\subsubsection*{Effects of data} 
\begin{figure}[t!]
\begin{center}
\includegraphics[trim=0in 2in 0in 2in, clip=true, width=3.5in]{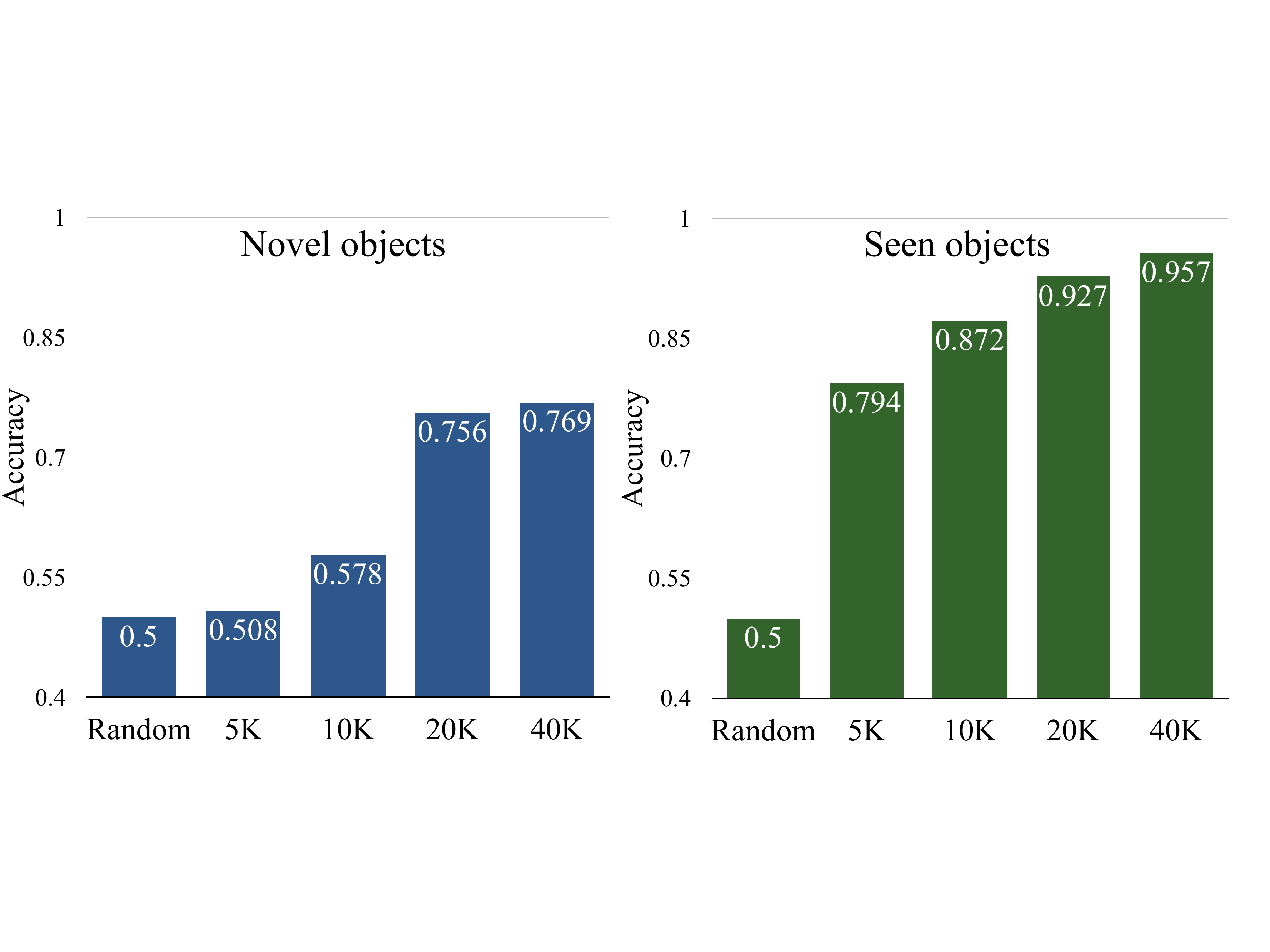}

\end{center}
\vspace{-0.1in}
\caption{Comparison of the performance of our learner over different training set sizes. Clear improvements in accuracy can be seen in both seen and unseen objects with increasing amounts of data.}
\vspace{-0.15in}
\label{fig:data_size_effect}
\end{figure}

It is seen in Fig.~\ref{fig:data_size_effect} that adding more data definitely helps in increasing accuracy. This increase is more prominent till about 20K data points after which the increase is small.

\subsubsection*{Effects of pretraining} An important question is how much boost does using pretrained network give. Our experiments suggest that this boost is significant: from accuracy of 64.6\% on scratch network to 76.9\% on pretrained networks. This means that visual features learnt from task of image classification~\cite{krizhevsky2012imagenet} aides the task of grasping objects.

\subsubsection*{Effects of multi-staged learning} After one stage of reinforcement, testing accuracy increases from 76.9\% to 79.3\%. This shows the effect of hard negatives in training where just 2K grasps improve more than from 20K random grasps. However this improvement in accuracy saturates to 79.5\% after 3 stages.

\subsubsection*{Effects of data aggregation} We notice that without aggregating data, and training the grasp model only with data from the current stage, accuracy falls from 76.9\% to 72.3\%. 

\subsection{Robot testing results}
\begin{figure}[t!]
\begin{center}
\includegraphics[width=3.3in]{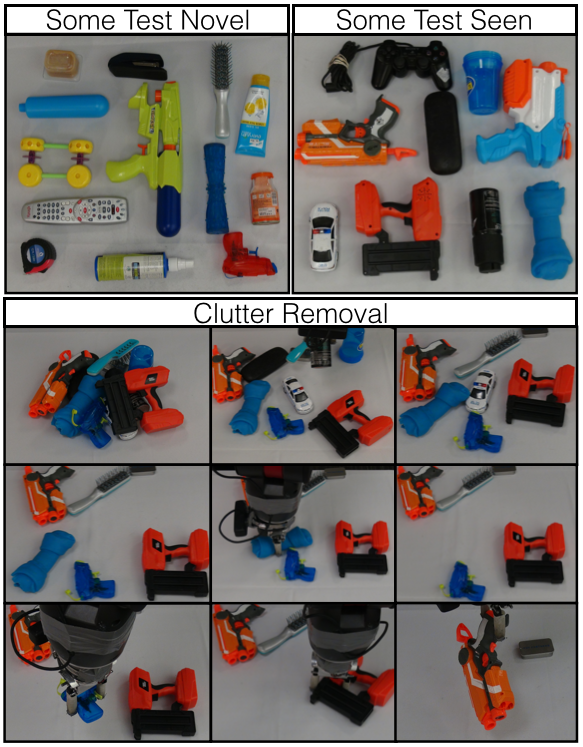}
\end{center}
\vspace{-0.13in}
\caption{Robot Testing Tasks: At test time we use both novel objects and training objects with different conditions. Clutter Removal is performed to show robustness of the grasping model}
\vspace{-0.21in}
\label{fig:novel_test_objects}
\end{figure}

Testing is performed over novel objects never seen by the robot before as well as some objects previously seen by the robot. Some of the novel objects can be seen in Fig.~\ref{fig:novel_test_objects}.


\noindent\textbf{Re-ranking Grasps:} One of the issues with Baxter is the precision of the arm. Therefore, to account for the imprecision, we sample the top 10 grasps and re-rank them based on neighborhood analysis: given an instance ($P_{topK}^i$,$\theta_{topK}^i$) of a top patch, we further sample 10 patches in the neighbourhood of $P_{topK}^i$. The average of the best angle scores for the neighbourhood patches is assigned as the new patch score $R_{topK}^i$ for the grasp configuration defined by  ($P_{topK}^i$,$\theta_{topK}^i$). The grasp configuration associated with the largest $R_{topK}^i$ is then executed. This step ensures that even if the execution of the grasp is off by a few millimeters, it should be successful.

\begin{figure*}[t!]
\begin{center}
\includegraphics[width=7in]{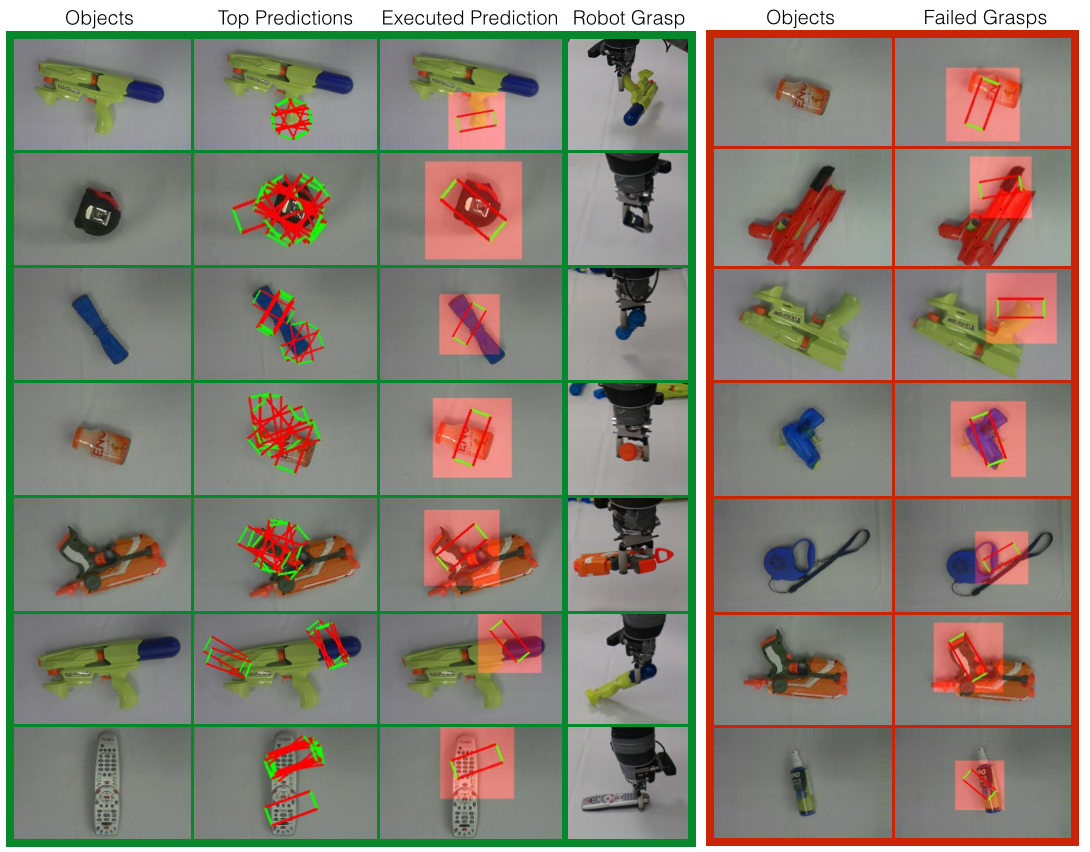}

\end{center}
\vspace{-0.15in}
\caption{Grasping Test Results: We demonstrate the grasping performance on both novel and seen objects. On the left (green border), we show the successful grasp executed by the Baxter robot. On the right (red border), we show some of the failure grasps. Overall, our robot shows 66\% grasp rate on novel objects and 73\% on seen objects.}
\vspace{-0.15in}
\label{fig:Testing_results_viz}
\end{figure*}

\noindent\textbf{Grasp Results:} We test the learnt grasp model both on novel objects and training objects under different pose conditions. A subset of the objects grasped along with failures in grasping can be seen in Fig.~\ref{fig:Testing_results_viz}. Note that some of the grasp such as the red gun in the second row are reasonable but still not successful due to the gripper size not being compatible with the width of the object. Other times even though the grasp is ``successful'', the object falls out due to slipping (green toy-gun in the third row). Finally, sometimes the impreciseness of Baxter also causes some failures in precision grasps.
Overall, of the 150 tries, Baxter grasps and raises novel objects to a height of 20 cm at a success rate of {\bf 66\%}. The grasping success rate for previously seen objects but in different conditions is {\bf 73\%}.

\noindent\textbf{Clutter Removal:} Since our data collection involves objects in clutter, we show that our model works not only on the objects in isolation but also on the challenging task of clutter removal~\cite{BoulariasBS15}. We attempted 5 tries at removing a clutter of 10 objects drawn from a mix of novel and previously seen objects. On an average, Baxter is successfully able to clear the clutter in 26 interactions.

\section{Conclusion}
We have presented a framework to self-supervise robot grasping task and shown that large-scale trial-error experiments are now possible. Unlike traditional grasping datasets/experiments which use a few hundred examples for training, we increase the training data 40x and collect 50K tries over 700 robot hours. Because of the scale of data collection, we show how we can train a high-capacity convolutional network for this task.  Even though we initialize using an Imagenet pre-trained network, our CNN has 18M new parameters to be trained. We compare our learnt grasp network to baselines and perform ablative studies for a deeper understanding on grasping. We finally show our network has good generalization performance with the grasp rate for novel objects being $66\%$. While this is just a small step in bringing big data to the field of robotics, we hope this will inspire the creation  of several other public datasets for robot interactions.

\section*{ACKNOWLEDGMENT}
This work was supported by ONR MURI N000141010934 and NSF IIS-1320083.

\bibliographystyle{unsrt}
\bibliography{references} 
\end{document}